\documentclass[journal,comsoc]{IEEEtran}

\usepackage[T1]{fontenc}% optional T1 font encoding

\usepackage{graphicx} %package to manage images
\usepackage{epstopdf}
\usepackage{footnote}
\usepackage{cite}
\usepackage{graphicx}

\usepackage{psfrag, epsfig}
\usepackage{subfig}

\usepackage{collcell}
\usepackage[table]{xcolor}
\usepackage[latin9]{inputenc}
\usepackage{hhline}
\usepackage{pgf}
\usepackage{multirow}
\usepackage[justification=centering]{caption}
\ifCLASSINFOpdf
\else
\fi
\usepackage{amsmath}
\usepackage[cmintegrals]{newtxmath}
\newcommand{\be}{\begin{equation}}
\newcommand{\ee}{\end{equation}}
\newcommand{\beqy}{\begin{eqnarray}}
\newcommand{\eeqy}{\end{eqnarray}}
\newcommand{\beqynn}{\begin{eqnarray*}}
\newcommand{\eeqynn}{\end{eqnarray*}}

\newcommand{\ba}{\begin{array}}
\newcommand{\ea}{\end{array}}
\newcommand{\bmx}{\begin{bmatrix}}
\newcommand{\emx}{\end{bmatrix}}
\newcommand{\bsmx}{\left[\begin{smallmatrix}}
\newcommand{\esmx}{\end{smallmatrix}\right]}
\newcommand{\bmxc}[1]{\left[\begin{array}{@{}#1@{}}}
\newcommand{\emxc}{\end{array}\right]}
\newcommand{\bt}[1]{\begin{tabular}{#1}}
\newcommand{\et}{\end{tabular}}
\newcommand{\bc}{\begin{center}}
\newcommand{\ec}{\end{center}}
\newcommand{\ben}{\begin{enumerate}}
\newcommand{\een}{\end{enumerate}}
\newcommand{\bi}{\begin{itemize}}
\newcommand{\ei}{\end{itemize}}

\newcommand{\x}{{\boldsymbol{x}}}

\begin{document}
\title{A Survey on Behaviour Recognition Using WiFi Channel State Information}

\author{Siamak Yousefi,~\IEEEmembership{Student Member,~IEEE,}
	Hirokazu Narui,~\IEEEmembership{}
	Sankalp Dayal,~\IEEEmembership{\\}
	Stefano Ermon,~\IEEEmembership{}
	Shahrokh~Valaee,~\IEEEmembership{Senior Member,~IEEE,}
	% <-this % stops a space
\thanks{S. Yousefi and S. Valaee are with the Department
of Electrical and Computer Engineering, University of Toronto, Ontario, Canada,
e-mail: siamak.yousefi@utoronto.ca, valaee@ece.utoronto.ca }% <-this % stops a space
\thanks{H. Narui, S. Dayal, and S. Ermon are with the Department of Computer Science, Stanford University, CA, USA.
e-mail: hirokaz2@stanford.edu, sankald@stanford.edu, ermon@cs.stanford.edu.}
% <-this % stops a space
\thanks{Manuscript received February, 2017. The work is supported in part by the Fund de Researche du Quebec-Nature et technologies (FQRNT) and FURUKAWA ELECTRIC Co., LTD.
The authors would like to thank Dr. Wei Wang of Nanjing University, China for the helpful discussions on CARM and sharing part of his code.}
}


\markboth{IEEE Communication Magazine, Draft}%
{Yousefi \MakeLowercase{\textit{et al.}}: A Survey of Human Activity Recognition Using WiFi CSI}



\maketitle

\begin{abstract}
In this article, we present a survey of recent advances in passive human behaviour recognition in indoor areas using the channel state information (CSI) of commercial WiFi systems.
Movement of human body causes a change in the wireless signal reflections, which results in variations in the CSI.
By analyzing the data streams of CSIs for different activities and comparing them against stored models, human behaviour can be recognized.
This is done by extracting features from CSI data streams and using machine learning techniques to build models and classifiers.
The techniques from the literature that are presented herein have great performances, however, instead of the machine learning techniques employed in these works, we propose to use deep learning techniques such as long-short term memory (LSTM) recurrent neural network (RNN), and show the improved performance.
We also discuss about different challenges such as environment change, frame rate selection, and multi-user scenario, and suggest possible directions for future work. 
\end{abstract}


% \fi
\IEEEpeerreviewmaketitle

\section{Background on Traditional Activity Recognition Systems}
\IEEEPARstart{H}{uman} activity recognition has gained tremendous attention in recent years due to numerous applications that aim to monitor the movement and behaviour of humans in indoor areas.
Applications such as health monitoring and fall detection for elderly people \cite{WiFall}, contextual awareness, activity recognition for energy efficiency in smart homes\cite{Wisee} and many other Internet of Things (IoT) based applications \cite{WiHear}.

In existing systems, the individual has to wear a device equipped with motion sensors such as gyroscope and accelerometer. 
The sensor data is processed locally on the wearable device or transmitted to a server for feature extraction and then supervised learning algorithms are used for classification. 
This type of monitoring is known as \emph{active} monitoring.
The performance of such system is shown to be around $90 \%$ for recognition of activities such as sleeping, sitting, standing, walking, and running \cite{politi2014human}. 

However, always wearing a device is cumbersome and may not be possible for many passive activity recognition applications, where the person may not be carrying any sensor or wireless device. 
While camera-based systems can be used for passive activity recognition, the line-of-sight (LOS) requirement is a major limitation for such systems. 
Furthermore, the camera-based approaches have privacy issues and cannot be employed in many environments.
Therefore, a passive monitoring system based on wireless signal, which does not violate the privacy of people, is desired.

Because of ubiquitous availability in indoor areas, recently, WiFi has been the focus of much research for activity recognition.
Such systems consist of a WiFi access point (AP) and one or several WiFi enabled devices located at different parts of the environment. 
When a person engages in an activity, body movement affects the wireless signals and changes the multi-path profile of the system.

\subsection{Techniques Based on Wi-Fi Signal Power}
The received signal strength (RSS) has been used successfully for active localization of wireless devices using WiFi fingerprinting techniques as summarized in \cite{Tahat_WiFi}.
% From_RSSI_to_CSI}.
The RSS has also been used as a metric for passive tracking of mobile objects \cite{RTI_Patwari}. 
When the person is located between the WiFi device and access point (AP), the signal will be attenuated and hence a different RSS is observed.
Although RSS is very simple to use and can be easily measured, it cannot capture the real changes in the signal due to the movement of the person. This is because the RSS is not a stable metric even when there is no dynamic change in the environment \cite{Adib_WiFi_Activity_Passive}.
%\cite{From_RSSI_to_CSI}.

\subsection{Techniques Requiring Modified WiFi Hardware}
To use some other metrics than RSS, in some systems, the WiFi system is modified so that extra information can be extracted from the signal.
The WiFi USRP software radio system is a modified WiFi hardware and has been used for  3D passive tracking in WiTrack \cite{Adib_WiFi_Activity_Passive}.
The idea is to measure the Doppler shift in the orthogonal frequency division multiplexing (OFDM) signals, caused due to movement of human body using a technique called frequency modulated carrier wave (FMCW). 
Since the Doppler shift is related to the distance, the location of the target can be estimated.
Using similar idea to WiTrack, in WiSee \cite{Wisee}, the USRP system is used to measure the Doppler shift in OFDM signals due to movement of human body.
The movement of the parts of the body toward the receiver causes positive Doppler shift, while moving the body parts away results in negative shift.
For instance for a gesture moving at $0.5$m/s, in a 5GHz system, the Doppler shift will be around $17$Hz \cite{Wisee}.
Therefore, such small Doppler shifts need to be detected in the system.  
In WiSee, the received signal is transformed into narrow-band pulses of few Hertz, and the WiSee receiver tracks the Doppler shift in the frequency of these pulses. 
After transforming the wide-band 802.11 to narrow-band pulses, the next steps in WiSee are as follows.

\subsubsection {Doppler Extraction}
To extract the Doppler information, WiSee computes the frequency-time Doppler profile by taking the FFT over samples in a window of half a second and then shifting the window by 5ms and continuing this process.
This technique is also known as short-time Fourier Transform (STFT), which was used in other techniques as well \cite{Spectrogram, CARM}.
Since the movement of human body generally has a speed of $0.25$m/s to $4$m/s, the Doppler shifts at 5 GHz is between 8Hz and 134Hz, hence only the FFT output in this frequency range is considered in WiSee.

\subsubsection{Segmentation}
The next step is to segment the STFT data to distinguish different patterns. For example, a gesture might consist of one segment with positive and negative Doppler shifts, or two or more segments, each of which has a positive and negative Doppler shift.
Detecting a segment is based on the energy detection over a small duration. If the energy is 3dB higher than the noise level, then the beginning of the segment is found and if it is less than 3dB, then the segment has ended.

\subsubsection{Classification}
The idea of classification is quite simple.
Each segment has three possibilities: only positive Doppler shifts, only negative Doppler shift, and segments with both positive and negative shifts, based on which three numbers are assigned to them. Thus each gesture is represented by a sequence of numbers. The classification task is to compare the obtained sequence with the one used during training.

WiSee also claims that the system can detect multiple moving targets and identify their activities using the idea that the reflections from each mobile target can be regarded as a signal from a wireless transmitter.
Therefore, using the idea used in multiple input multiple output (MIMO) receivers, the reflected signals due to different people moving in the area can be separated.
The problem is to find the weight matrix that when multiplied with the Doppler energy corresponding to each segment of each antenna, maximizes the Doppler of each segment.
To this end, iterative algorithms have been employed.

In contrast to techniques such as WiSee that requires specialized USRP software radios, there has been several efforts to employ commercial WiFi APs without the need to modify the WiFi system.
To represent the dynamic changes in the environment due to movement of human body, recently other metrics have been employed, such as channel state information (CSI), which will be described in more details below.

\section{WiFi Channel State Information}

%\section{An Overview of IEEE 802.11 Systems}\label{Sec:Overview}
\subsection{CSI of WiFi System} 
The wireless devices with IEEE 802.11n/ac standards are using multiple input multiple output (MIMO) system.
By using MIMO technology, it is possible to increase the diversity gain, array gain, and multiplexing gain, while reducing the co-channel interference \cite{80211N}.
The modulation used in IEEE 802.11 is OFDM where the bandwidth is shared among multiple orthogonal subcarriers.
Due to the small bandwidth, the fading that each subcarrier faces is modeled as flat fading.
Therefore, using OFDM, the small scale fading property of the channel can be mitigated.

Let $M_T$ denote the number of transmit antennas at the device, and $M_R$ the number of receive antennas at the AP.
The MIMO system at any time instant can be modeled as 
$\mathbf{y}_i = \mathbf{H}_i \mathbf{x}_i + \mathbf{n}_i $, for $i \in \{1, \ldots ,S\}$
where $S$ is the number of OFDM subcarriers, and $\mathbf{x}_i \in \mathbb{R}^{M_T}$ and $\mathbf{y}_i \in \mathbb{R}^{M_R}$ represent the transmit and received signal vectors for the $i$-th subcarrier, respectively, and $\mathbf{n}_i $ is the noise vector.
The channel matrix for $i$-th subcarrier $\mathbf{H}_i$, which consists of complex values, can be estimated by dividing the output signal with a known sequence of input also known as pilot.
The channel matrix is also known as the CSI, as it shows how the input symbol is affected by the channel to reach at the receiver.
In OFDM systems, each subcarrier faces a narrow-band fading channel, and by obtaining the CSI for each subcarrier, there will be diversity in the observed channel dynamics. This is the main advantage of using CSI compared to RSS, in which the changes are averaged out over all the WiFi bandwidth and hence cannot capture the change at certain frequencies.
In some commercial network interface cards (NICs), such as Intel NIC 5300 the CSI can be collected using the tool provided in \cite{Halperin_csitool}.

\subsection{Limitations and Errors of WiFi Systems}
The amplitude of CSI is generally a reliable metric to use for feature extraction and classification, although it can change with transmission power, and transmission rate adaptation.
As will be discussed later, by using filtering techniques, the burst noise can be reduced \cite{CARM}.
However, in contrast to amplitude, the phase of WiFi system is affected by several sources of error such as carrier frequency offset (CFO) and sampling frequency offset (SFO).
The CFO exists due to the difference in central frequencies (lack of synchronization) between the transmitter and receiver clocks.
The CFO for a period of $50 \mu s$ of 5GHz WiFi band can be as large as $80$KHz, leading to phase change of $8 \pi$.
Therefore, the phase changes due to the movement of the body, which is generally smaller than $0.5 \pi$, is not observable from phase change.
The other source of error, SFO, is generated by the receiver analog to digital converter (ADC).
The SFO is also varying by subcarrier index, therefore, each subcarrier faces a different error.

\begin{figure*}[htbp]
\centering
\subfloat[CSI amplitude]{
\includegraphics[width=50mm,height=35mm]{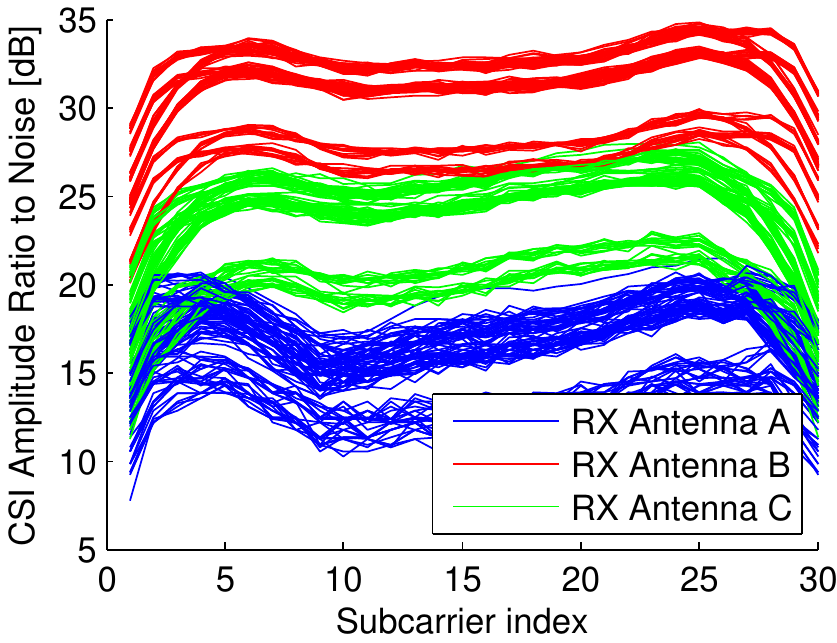}
} \quad
%\vspace{0.1cm}
\subfloat[CSI phase]{
\includegraphics[width=50mm,height=35mm]{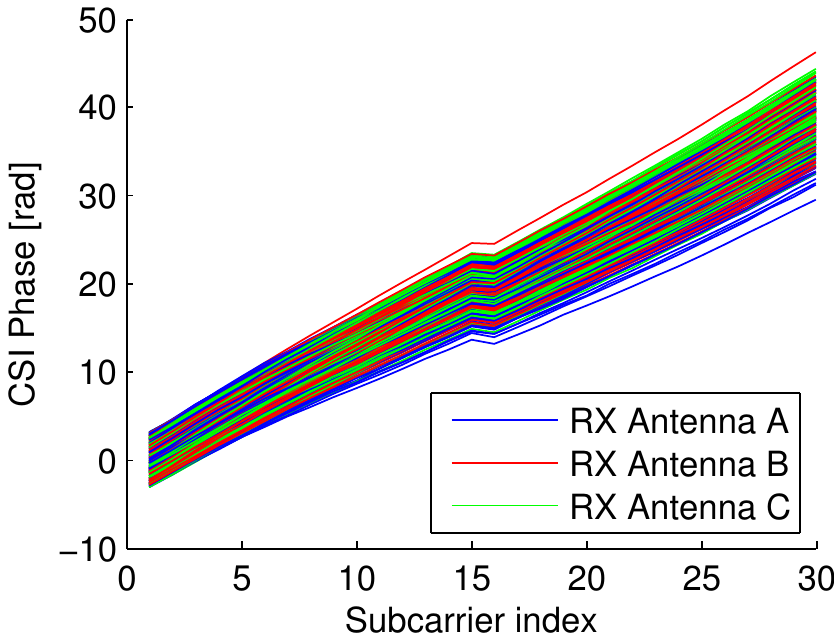}
} \quad
%\vspace{0.1cm}
\subfloat[Sanitized CSI phase]{
\includegraphics[width=50mm,height= 35mm]{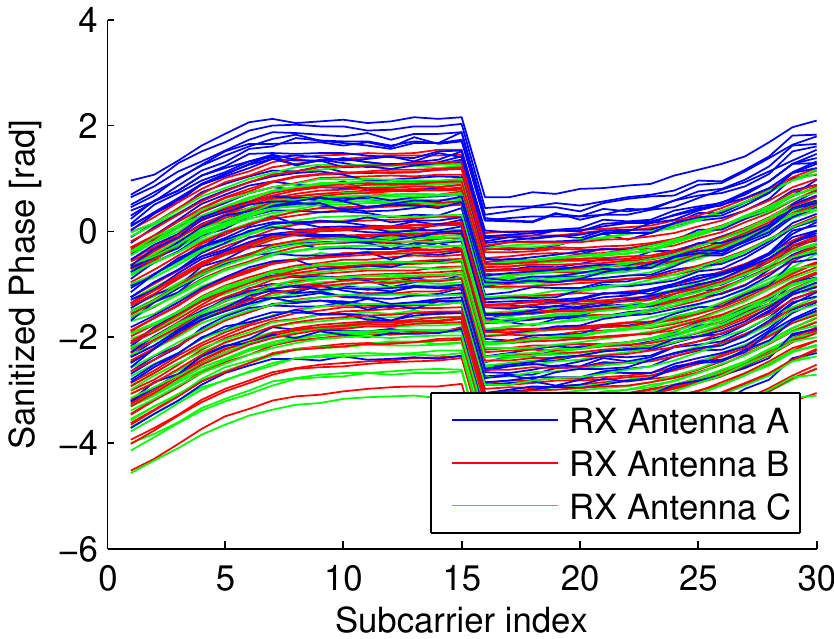}
}
\caption{CSI measured in LOS condition for three antennas as a function of subcarrier index: (a) amplitude of CSI, (b) phase of CSI, (c) Sanitized phase of CSI.}
\label{fig:CSI_Subcarrier}
\end{figure*}

Due to the unknown CFO and SFO, using the raw phase information may not be useful.
However, a linear transformation is proposed in \cite{Phase_Sanitization}, such that the CFO and SFO can be removed from the calibrated phase.
This process is also known as phase sanitization.
In Fig. \ref{fig:CSI_Subcarrier}, the CSI amplitude, CSI phase and sanitized CSI phase versus the subcarrier index are plotted for a scenario where the WiFi transmitter and receiver are located in the vicinity of each other in LOS condition.
As observed, the CSI amplitude is relatively stable but forms some clusters as mentioned in \cite{Phase_Sanitization}.
The raw phase increases with subcarrier index since the SFO grows with subcarrier index, as illustrated in Fig. \ref{fig:CSI_Subcarrier}-(b).
After phase sanitization, the change of phase due to SFO will be reduced as observed in Fig. \ref{fig:CSI_Subcarrier}-(c).

\subsection{Effect of Human Motion on Wireless Channel}
\begin{figure}[htbp]
\centering
\subfloat[CSI amplitude for three antennas as a function of time.]{
\includegraphics[clip, trim=4cm -1.3cm 5.0cm 0.1cm, scale=0.35,angle=90]{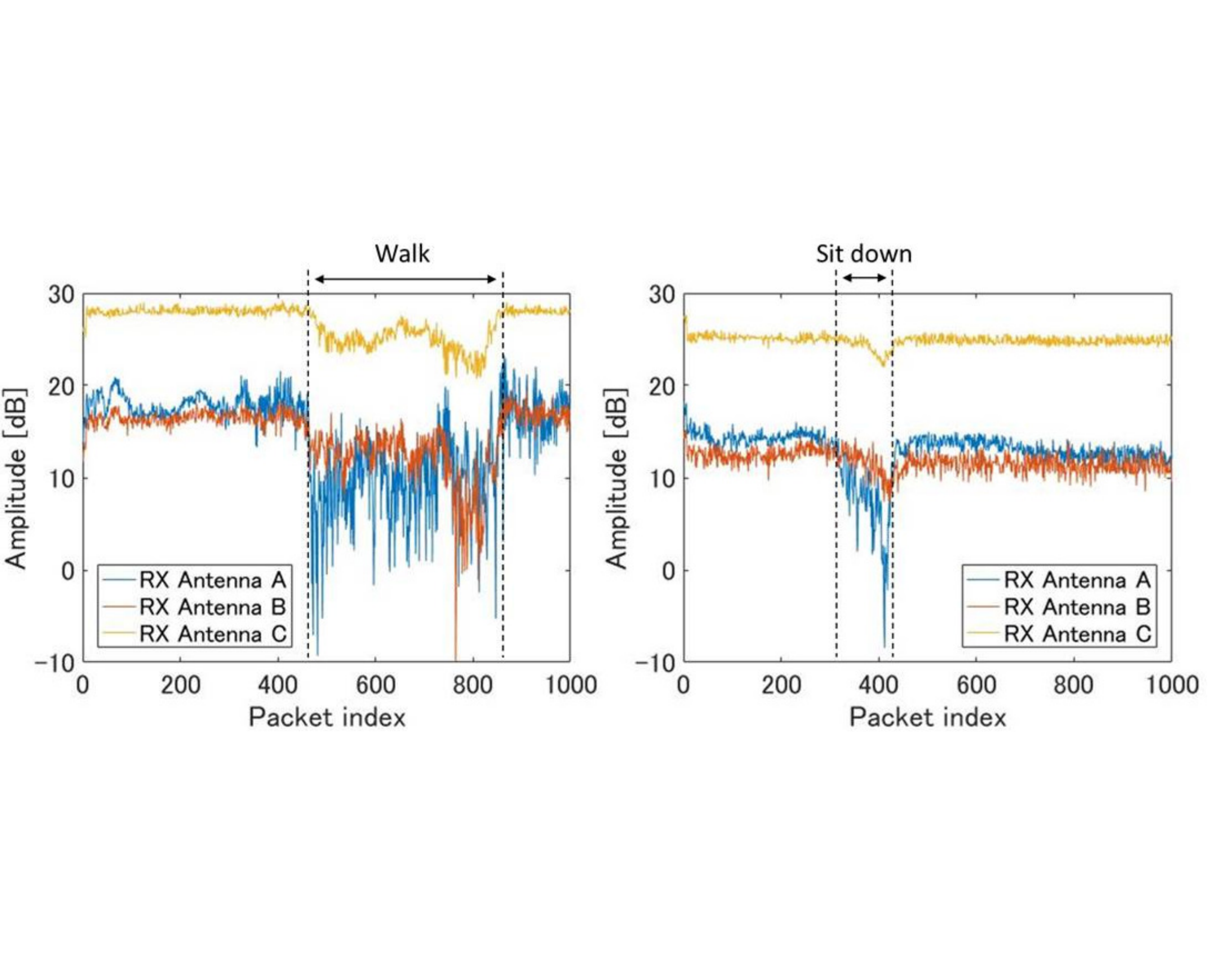}
}
\vspace{0.1cm}
\subfloat[CSI phase for three antennas as a function of time.]{
\includegraphics[clip, trim=4cm -1.3cm 5.0cm 0.1cm, scale=0.35,angle=90]{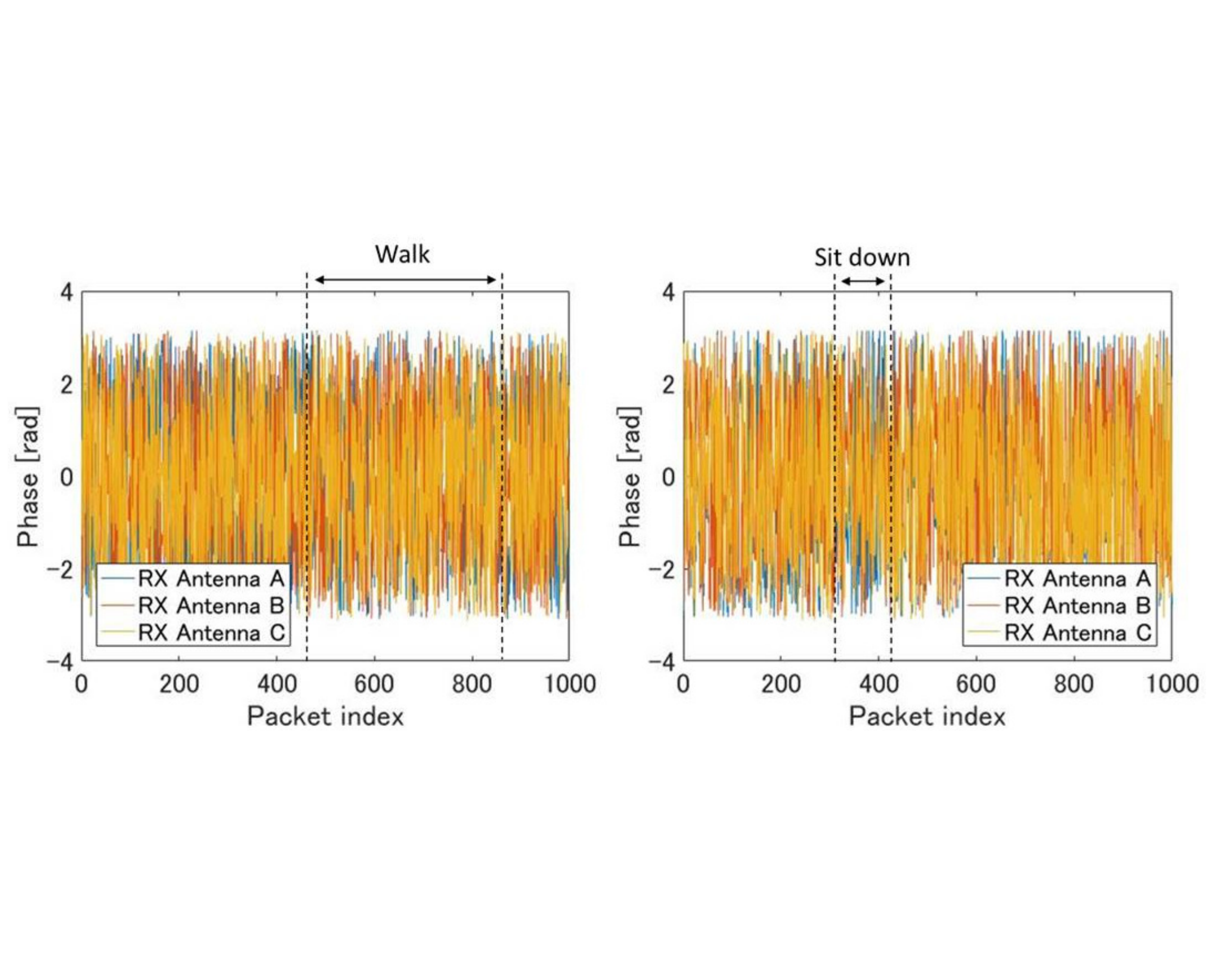}
}
\vspace{0.1cm}
\subfloat[Sanitized CSI phase for three antennas as a function of time.]{
\includegraphics[clip, trim=4cm -1.3cm 5.0cm 0.1cm, scale=0.35,angle=90]{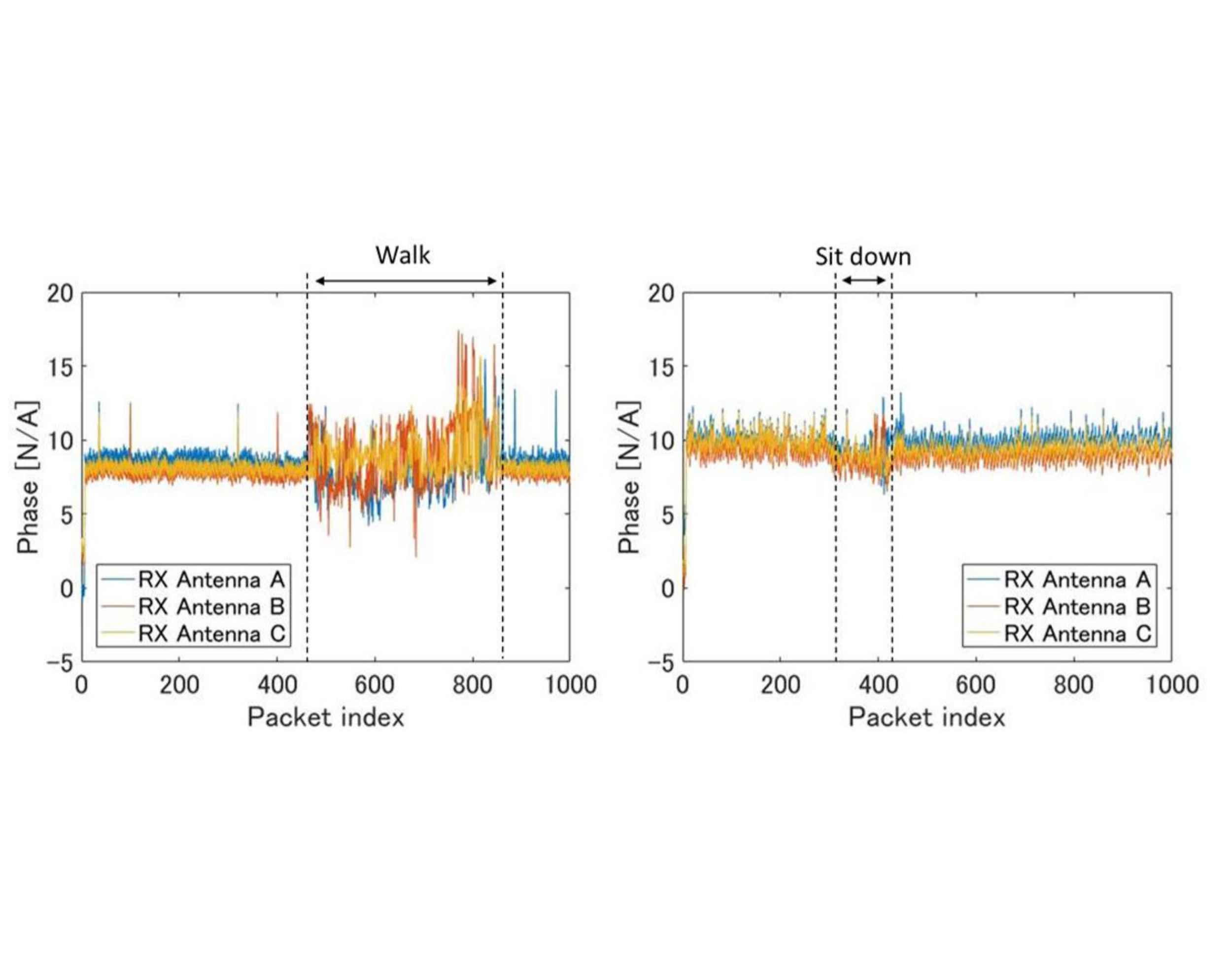}
}
\caption{CSI Changes Under Human Motion}
\label{fig:CSI}
\end{figure}

The movement of the humans and objects change the multipath characteristic of the wireless channel and hence the estimated channel will have a different amplitude and phase.
The CSI amplitude for one subcarrier and all the antennas, related to a person walking and sitting down between a WiFi transmitter and receiver, is illustrated in Fig. \ref{fig:CSI}-(a).
The person is stationary for the first 400 packets but then starts walking or sitting down.
As observed, when the person is not moving, the CSI amplitudes for all antennas are relatively stable, however, when the activity starts, the CSIs start changing drastically.
The walking period is longer than sitting in this experiment because when the person sits down he remains stationary.

The received phase, is very distorted due to the CFO and SFO, as mentioned earlier. 
This can be observed in Fig. \ref{fig:CSI}-(b).
However, using the phase sanitization technique, the effect of errors in phase can be eliminated.
The calibrated phase can be observed in Fig. \ref{fig:CSI}-(c).

\section{Wi-Fi CSI-based Behaviuor Recognition} \label{Sec:Past_Techniques}

\begin{figure*}[htbp]
\centering
\includegraphics[trim =0 0 0 0 0, clip,height=3in,width=6.0in]{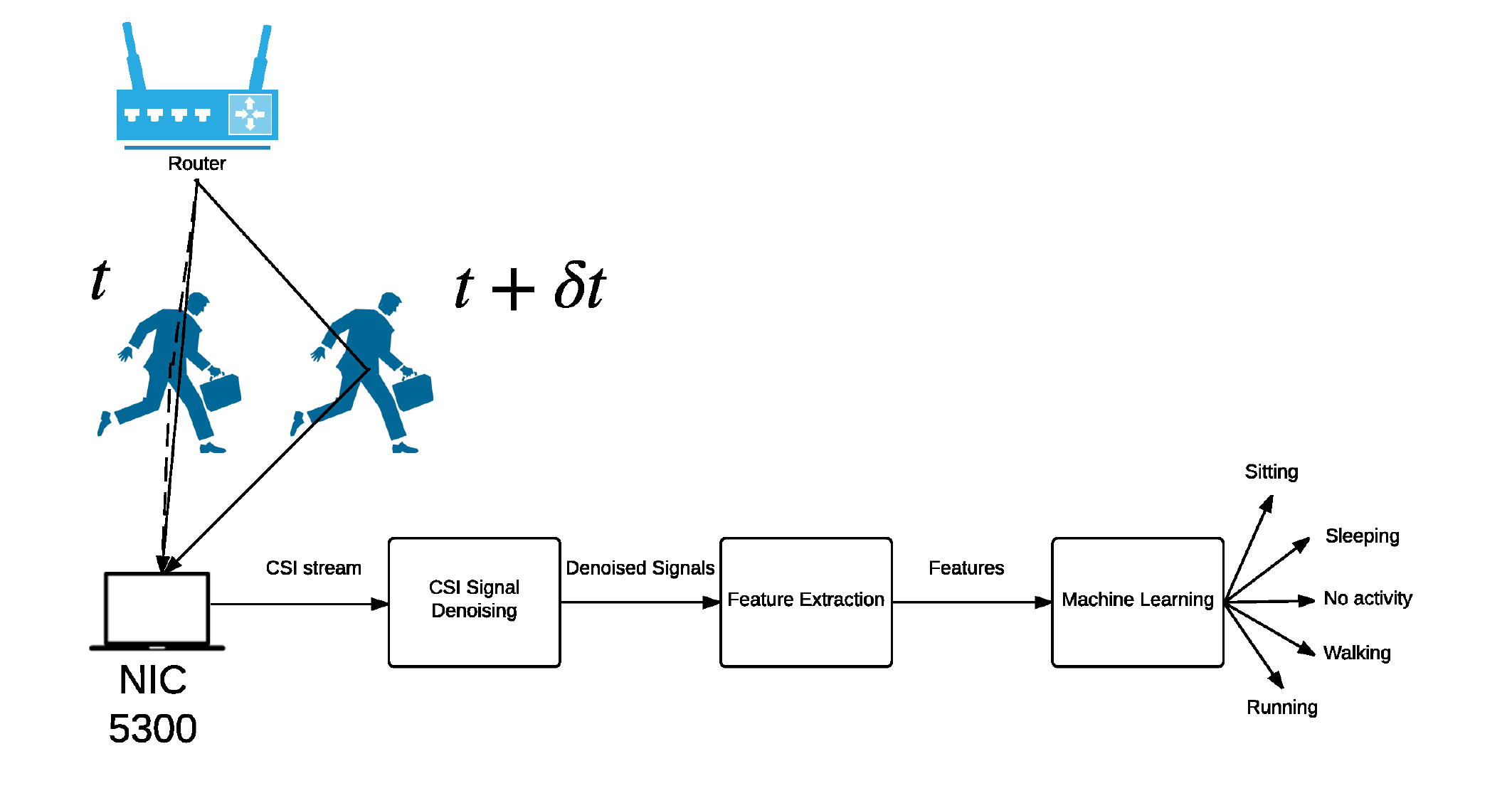}
\caption{The scheme of common activity recognition techniques. A person is moving in the area between the router and WiFi device from time $t$ to time $t+\delta t$.}
\label{fig:Activity_Recognition_Scheme}
\end{figure*}

In this section, we provide a summary of the techniques using commercial WiFi NICs.
The general diagram of activity recognition systems using WiFi CSI is illustrated in Fig. \ref{fig:Activity_Recognition_Scheme}.

\subsection{Histogram-based Techniques}
One of these technique is E-Eyes \cite{E_Eyes_WiFi_Activity_Passive}, in which the CSI histograms are used as fingerprints in a database.
In test phase, by comparing the histogram of the obtained CSI with the database, the closest one is found and hence the activity can be recognized.
The preprocessing steps is to do low-pass filtering and modulation and coding scheme (MCS) index filtering. 
The former is necessary to remove the high frequency noise, which may not be due to the human movement, and the latter is needed to reduce the unstable wireless channel variations. 
Although the performance of this technique is good and its computational cost is low, the histogram technique is sensitive to environment changes and hence may not perform well for varying environments.

\subsection{CARM}
Recently, other techniques have been proposed such as WiHear \cite{WiHear}, CARM \cite{CARM}, and \cite{Kamran_Keystroke_Recognition}.
In WiHear, directional antennas are used to capture CSI variations caused by the movement of mouth.
The performance of WiHear is good, however, the application is only to monitor the spoken words.
In \cite{Kamran_Keystroke_Recognition}, the authors use advanced feature extraction and machine learning techniques for recognition of typed word on a keyboard.
The idea is similar to the idea in CARM \cite{CARM}, which will be described in more detail below.

\subsubsection{CSI De-noising}
The CSI is noisy and may not show distinctive features for different activities.
Therefore, it is necessary to first filter-out the noise and then extract some features to be used for classification using machine learning techniques.
There are different methods for filtering the noise such as Butterworth low-pass filters \cite{CARM}.
However, due to the existence of burst and impulse noises in CSI, which have high bandwidths, the low-pass filter cannot yield a smooth CSI stream \cite{CARM}. 

It has been shown that there are better techniques for this purpose such as principal component analysis (PCA) de-noising \cite{CARM}.
The PCA is a technique for dimensionality reduction of a large dimension system using the idea that most of the information about the signal is concentrated over some of the features.
In CARM, the first principal component is discarded to reduce the noise, and the next five ones are employed for feature extraction.
By removing the first principal component, the information due to the dynamic reflection coming from mobile target is not lost because it is also captured in other principal components.
After PCA de-noising of CSI data, some features are extracted from it so it can be used for classification. The feature extraction will be discussed below.

\subsubsection{Feature Extraction}
\begin{figure*}[htbp]
\centering
\subfloat[]{
\includegraphics[width=60mm,height=45mm]{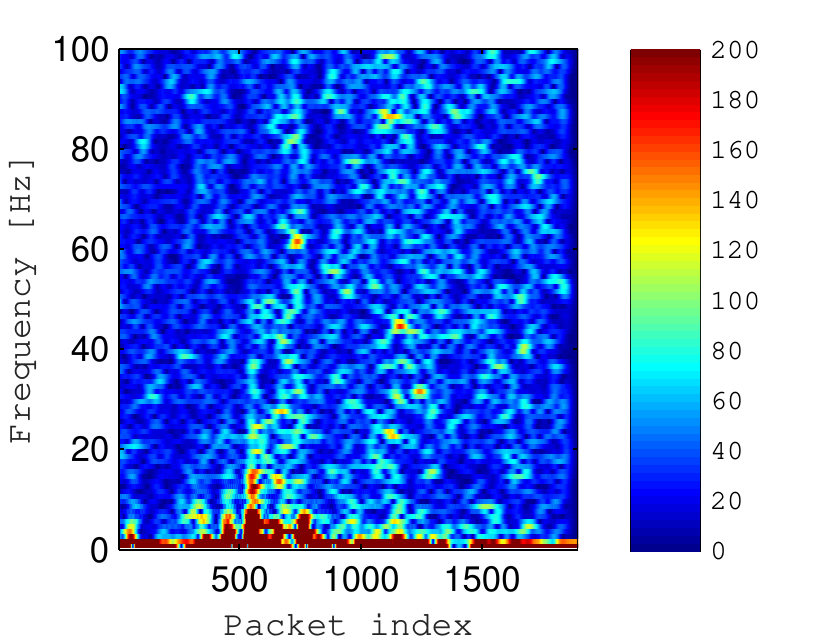}
}~
\subfloat[]{
\includegraphics[width=60mm,height=45mm]{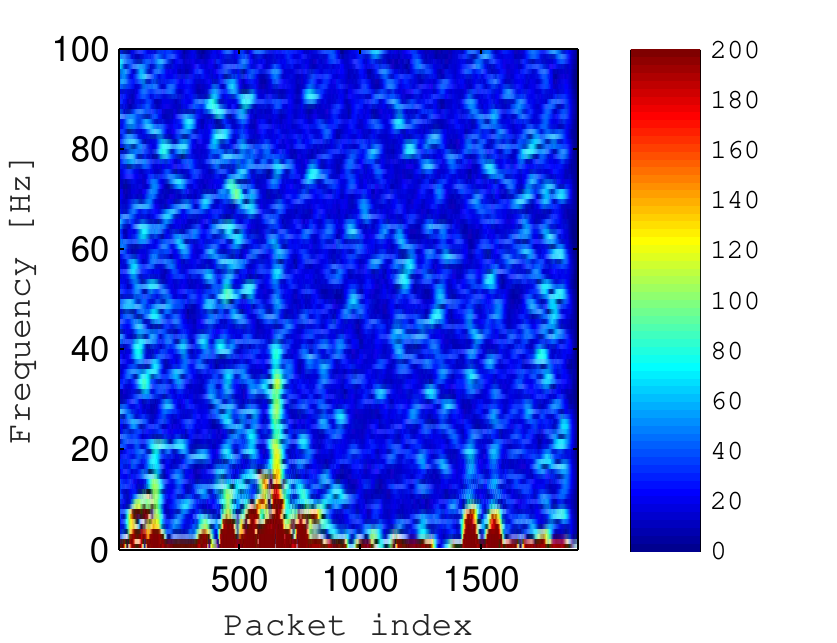}
}~
\subfloat[]{
\includegraphics[width=60mm,height=45mm]{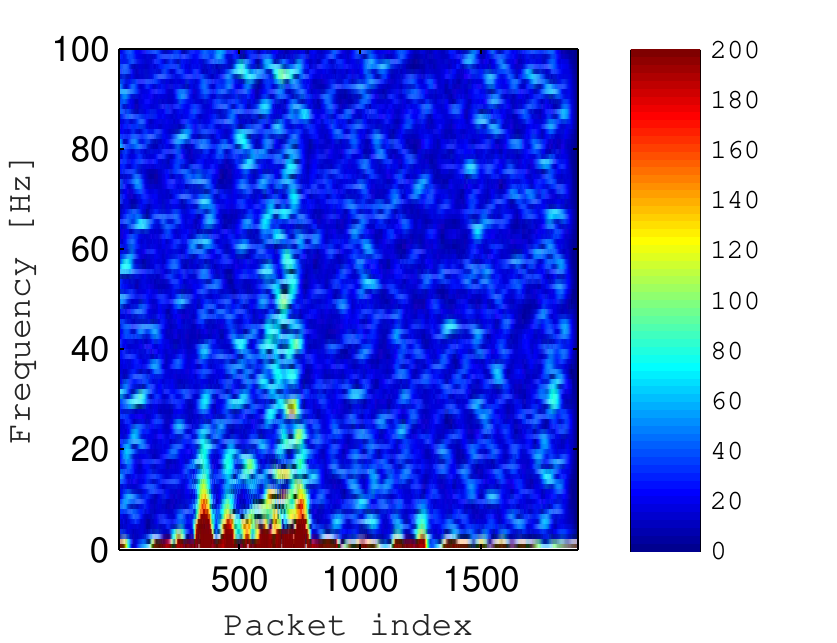}
}\\
\subfloat[]{
\includegraphics[width=60mm,height=45mm]{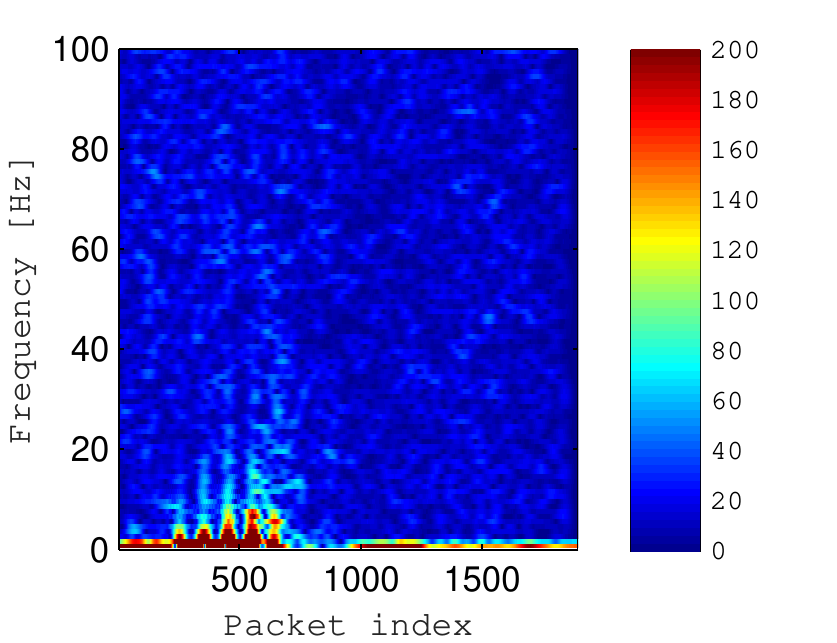}
}~
\subfloat[]{
\includegraphics[width=60mm,height=45mm]{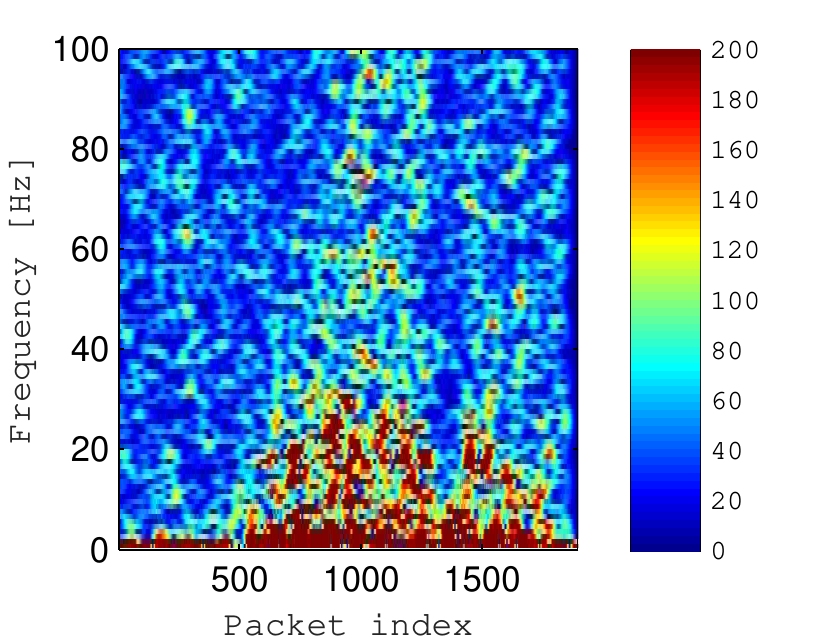}
}~
\subfloat[]{
\includegraphics[width=60mm,height=45mm]{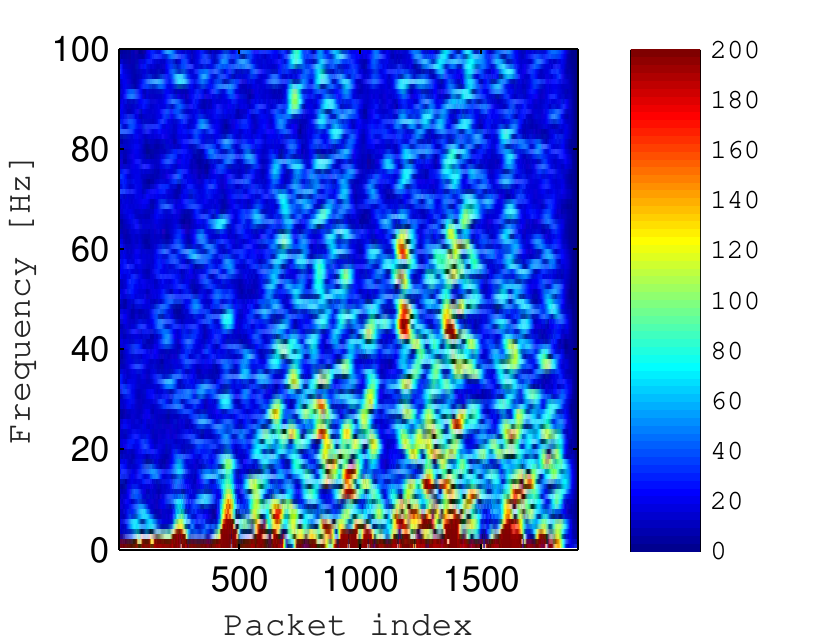}
}
\caption{The spectrogram of one subcarrier's CSI amplitude for different activities: (a) standing up, (b) sitting down, (c) laying down, (d) falling, (e) walking, (f) running.}
\label{fig:Spectrogram}
\end{figure*}

One way to extract features from a signal is to transform it to another domain, such as frequency domain.
The fast Fourier transform (FFT), which is an efficient implementation of discrete Fourier transform (DFT) can be used for this purpose.
To this end, a window size of certain number of CSI samples is selected and then the FFT is applied on each segment by sliding the window.
This technique is known as short-time Fourier transform (STFT), which can detect the frequency changes of a signal over time.
The STFT was used in radar for detection of the movement of torso and legs in \cite{Spectrogram}.
In Fig. \ref{fig:Spectrogram}, the STFT (spectrogram) of CSI for different activities is shown for CSI data collected at 1KHz rate.
As observed in Fig. \ref{fig:Spectrogram}, the activities that involve drastic movements such as walking and running show high energy in high frequencies in the spectrogram.

In \cite{CARM, Kamran_Keystroke_Recognition, WiHear}, DWT is employed to extract features from CSI as a function of time.
The DWT provides high time resolution for activities with high frequencies and high frequency resolution for activities with low speeds.
Each level of DWT represents a frequency range, where the lower levels contain higher frequency information while higher levels contain lower frequencies.
The advantage of DWT to STFT as mentioned in \cite{CARM} are:
\begin{itemize}
\item DWT can provide a nice trade-off in time and frequency domain,
\item DWT reduces the size of the data as well, so it becomes suitable for machine learning algorithms.
\end{itemize}

In CARM, a 12 level DWT is employed to decompose the five principal components (after removing the first principal component).
Then the five values of the DWT are averaged.
For every 200ms, CARM extracts a 27 dimensional feature vector including three sets of features:
\begin{itemize}
\item the energy in each wavelet level, representing the intensity of movements with different speeds,
\item the difference in each level between consecutive 200ms intervals,
\item the torso and leg speeds estimated using the Doppler radar technique \cite{Spectrogram}.
\end{itemize}
These features are used as the input to the classification algorithm described below.

\subsubsection{Machine Learning for Classification}
Different machine learning techniques can be used for multi-class classification based on certain features that are extracted.
Some of the popular classification techniques are logistic regression, support vector machines (SVM), hidden Markov model (HMM), and deep learning.  
Since the activity data is in a sequence, CARM uses HMM and it is shown that satisfactory results can be obtained.

\subsection{Using Deep Learning for Behaviour Recognition}
The problem of activity recognition is somewhat similar to the speech recognition process, where traditionally HMM has been used for classification.
However, deep recurrent neural network (RNN) has been considered as a counterpart of HMM.
Training RNN is difficult as it suffers from vanishing or exploding gradient problem, however, it has been shown in \cite{Speech_RNN} that using the long short term memory (LSTM) extension of RNN, the best accuracy for speech recognition so far can be achieved.
Therefore, we propose using LSTM for activity recognition rather than other conventional machine learning techniques, such as HMM, although feature extraction is not done similar to CARM. 
Using LSTM has two advantages. First, the LSTM can extract the features automatically, in other words, there is no necessity to pre-process the data.
Second, LSTM can hold temporal state information of the activity, i.e., LSTM has the potential to distinguish similar activities like "Lay down" and "Fall".
Since "Lay down" consist of "Sit down" and "Fall", the memory of LSTM can help in recognition of these activities.

\section{Evaluation of Different Methods}
%\label{Sec:Experiments}
In this section, we implement different methods as well as our proposed method and show the performance of each one. 

\subsection{Measurement Setup}
We do the experiments in an indoor office area where the Tx and Rx are located 3 meters apart in LOS condition. The Rx is equipped with a commercial Intel 5300 NIC, with sampling rate of 1KHz.
A person starts moving and doing an activity within a period of 20 seconds in LOS condition, while in the beginning and at the end of the time period the person remains stationary. 
We also record videos of activities so we can label the data.
Our data set includes 6 persons, 6 activities, denoted as "Lay down, Fall, Walk, Run, Sit down, Stand up", and 20 trials for each one.

\subsection{Evaluating Machine Learning Techniques}
We apply the PCA on the CSI amplitude, and then use STFT to extract features in the frequency domain for every 100 ms.
We only use the first 25 frequency components out of 128 frequency bins as most of the energy of activities is in lower frequencies, and in this way, the feature vector does not become sparse.

First, we use random forest with 100 trees for classification of activities.
To have a feature vector that contains enough information about an activity, the modified STFT bins are stacked together in a vector for every 2 seconds of activity, hence every feature vector will be of length 1000.
We also implemented other techniques such as support vector machine, logistic regression, and decision tree, however, random forest outperformed these techniques.
The confusion matrix for random forest is shown in Table I-(a) and as observed, decent performance can be obtained for some of the activities, but not for activities such as "Lay down", "Sit down" and "Stand up".

We also apply HMM on the extracted features using STFT and use the MATLAB toolbox for HMM training.
Note that HMM is also used in CARM, however, DWT and the technique in \cite{Spectrogram} are used for feature extraction.
The result is shown in Table I-(b) and improved accuracy compared to random forest can be observed, although with higher computation time needed for training.
Although the performance of HMM is good, especially for "Walk" and "Run", it sometimes miss-classifies "Stand up" with "Sit down" or "Lay down".

We evaluate the performance of LSTM using Tensorflow in Python.
The input feature vector is the raw CSI amplitude data, which is a 90 dimensional vector (3 antennas and 30 sub-carriers). 
The LSTM approach is different from conventional approaches in the sense that it does not use PCA and STFT and can extract features from CSI directly.
The number of hidden units is chosen to be 200 where we consider only one hidden layer.
For numerical minimization of cross entropy, we use the Stochastic Gradient Descent (SGD) with batch size 200 and learning rate of $10^{-4}$. 
%We evaluate the accuracy using 10-fold cross validation.
Our result is shown in Table I-(c), where the accuracy is over 75\% for all activities.
One of the drawbacks of using LSTM in this way is the long training time compared to HMM, however, using deep learning packages such as Tensorflow one can also use GPUs and speed up the training.
Once the LSTM is trained the test can be done quickly.

% Confusion matrix
\def\colorModel{hsb} %You can use rgb or hsb

\newcommand\ColCell[1]{
  \pgfmathparse{#1<.5?1:0}  %Threshold for changing the font color into the cells
    \ifnum\pgfmathresult=0\relax\color{white}\fi
  \pgfmathsetmacro\compA{0}      %Component R or H
  \pgfmathsetmacro\compB{#1-0} %Component G or S
  \pgfmathsetmacro\compC{1}      %Component B or B
  \edef\x{\noexpand\centering\noexpand\cellcolor[\colorModel]{\compA,\compB,\compC}}\x #1
  } 
\newcolumntype{E}{>{\collectcell\ColCell}m{0.2cm}<{\endcollectcell}}  %Cell width
\newcommand*\rot{\rotatebox{90}}
\newcolumntype{E}{>{\collectcell\ColCell}c<{\endcollectcell}}

\newcommand\items{7} %Number of classes
\arrayrulecolor{white} %Table line colors
\begin{table}[h]\caption{Confusion matrix}

\centering

\subfloat[Random Forest]{
	
	\scalebox{0.8}{
		
		\noindent\begin{tabular}{cc*{\items}{|E}|}
			
			\multicolumn{1}{c}{}  &\multicolumn{\items}{c}{Predicted}  \\ \hhline{~*\items{|-}|}
			
			\multicolumn{1}{c}{} &\multicolumn{1}{c}{} &\multicolumn{1}{c}{\rot{Lay down}} &\multicolumn{1}{c}{\rot{Fall}} &\multicolumn{1}{c}{\rot{Walk}} &\multicolumn{1}{c}{\rot{Run}} &\multicolumn{1}{c}{\rot{Sit down}} &\multicolumn{1}{c}{\rot{Stand up}}  \\ \hhline{~*\items{|-}|}
			
			\multirow{\items}{*}{\rotatebox{90}{Actual}}
			
			&Lay down & 0.53 & 0.03 & 0.0 & 0.0 & 0.23 & 0.21  \\ \hhline{~*\items{|-}|}
			
			&Fall & 0.15 & 0.60 & 0.03 & 0.07 & 0.1 & 0.05  \\ \hhline{~*\items{|-}|}
			
			&Walk & 0.04 & 0.05 & 0.81 & 0.07 & 0.01 & 0.01  \\ \hhline{~*\items{|-}|}
			
			&Run & 0.01 & 0.03 & 0.05 &  0.88 & 0.01 & 0.01  \\ \hhline{~*\items{|-}|}
			
%			&Pick up & 0.05 & 0.06 & 0.06 & 0.75 & 0.03 & 0.04 & 0.01 \\ \hhline{~*\items{|-}|}
			
			&Sit down & 0.15 & 0.03 & 0.02  & 0.04 & 0.49 & 0.26 \\ \hhline{~*\items{|-}|}
			
			&Stand up & 0.10 & 0.03 & 0.02 & 0.06 &  0.20 & 0.57   \\ \hhline{~*\items{|-}|}

			\label{DT}
			
		\end{tabular}
		
}}

\subfloat[Hidden Markov Model]{
	
	\scalebox{0.8}{
		
		\noindent\begin{tabular}{cc*{\items}{|E}|}
			
			\multicolumn{1}{c}{}  &\multicolumn{\items}{c}{Predicted} \\ \hhline{~*\items{|-}|}
			
			\multicolumn{1}{c}{} &\multicolumn{1}{c}{} &\multicolumn{1}{c}{\rot{Lay down}} &\multicolumn{1}{c}{\rot{Fall}} &\multicolumn{1}{c}{\rot{Walk}} &\multicolumn{1}{c}{\rot{Run}}  &\multicolumn{1}{c}{\rot{Sit down}} &\multicolumn{1}{c}{\rot{Stand up}}  \\ \hhline{~*\items{|-}|}
			
			\multirow{\items}{*}{\rotatebox{90}{Actual}}
			
			&Lay down & 0.52 & 0.08 & 0.08&  0.16 & 0.12 & 0.04 \\ \hhline{~*\items{|-}|}
			
			&Fall & 0.08 & 0.72 & 0.0 &  0.0 & 0.2 & 0.0 \\ \hhline{~*\items{|-}|}
			
			&Walk & 0.0 & 0.04 & 0.92 & 0.04 & 0.0 & 0.0 \\ \hhline{~*\items{|-}|}
			
%			&Pick up & 0.05 & 0.06 & 0.06 & 0.75 & 0.03 & 0.04 & 0.01 \\ \hhline{~*\items{|-}|}
			
			&Run & 0.0 & 0.0 & 0.04 & 0.96 & 0.0 & 0.0 \\ \hhline{~*\items{|-}|}
			
			&Sit down & 0.0 & 0.04 &  0.0 & 0.0 & 0.76 & 0.20 \\ \hhline{~*\items{|-}|}
			
			&Stand up & 0.16 & 0.04 & 0.0 &  0.0 & 0.28 & 0.52 \\ \hhline{~*\items{|-}|}
			
			\label{HMM}
			
		\end{tabular}
		
}}

\subfloat[Long Short Term Memory]{

	\scalebox{0.8}{

		\noindent\begin{tabular}{cc*{\items}{|E}|}

			\multicolumn{1}{c}{}  &\multicolumn{\items}{c}{Predicted} \\ \hhline{~*\items{|-}|}

			\multicolumn{1}{c}{} &\multicolumn{1}{c}{} &\multicolumn{1}{c}{\rot{Lay down}} &\multicolumn{1}{c}{\rot{Fall}} &\multicolumn{1}{c}{\rot{Walk}} &\multicolumn{1}{c}{\rot{Run}}  &\multicolumn{1}{c}{\rot{Sit down}} &\multicolumn{1}{c}{\rot{Stand up}}  \\ \hhline{~*\items{|-}|}

			\multirow{\items}{*}{\rotatebox{90}{Actual}}

% 1 kHz result
			&Lay down & 0.95 & 0.01 & 0.01  & 0.01 & 0.00  & 0.02 \\ \hhline{~*\items{|-}|}

			&Fall & 0.01 & 0.94 & 0.05   & 0.00 & 0.00 & 0.00 \\ \hhline{~*\items{|-}|}

			&Walk & 0.00 & 0.01 & 0.93  & 0.04 & 0.01 & 0.01  \\ \hhline{~*\items{|-}|}

			&Run  & 0.00 & 0.00 & 0.02  & 0.97 & 0.01  & 0.00 \\ \hhline{~*\items{|-}|}

			&Sit down  & 0.03 & 0.01 & 0.05  & 0.02 & 0.81 & 0.07  \\ \hhline{~*\items{|-}|}

			&Stand up  & 0.01 & 0.00 & 0.03  & 0.05 &0.07 & 0.83  \\ \hhline{~*\items{|-}|}

			\label{LSTM}

		\end{tabular}

}}

\end{table}

\section{Discussions} \label{Sec:Discussions}

\subsection{Effect of Environment Change on the Performance}
The CSI characteristics are not the same for different environments and different people.
%%%%% 
There are different techniques to reduce the influence of environments \cite{CARM}. 
For instance, after using PCA, the first component includes mainly the CSI information due to stationary objects \cite{CARM}. 
By discarding the first principal component, the information due to mobile target is mainly captured.
Therefore, using this technique relatively similar features can be obtained for different environments.
Other techniques such as STFT and DWT represent the speed of change in the multi-paths which is related to the speed of movement of human body parts. Although the same activities in different environments result in very different CSI characteristics, due to similarity in the change of signal reflections, similar features can be obtained for different environments and people using STFT or DWT \cite{CARM}.

\subsection{Effect of Wi-Fi transmission rate on the performance}
In order for the CSI to show noticeable changes due to the movement, the rate of transmission should be high enough (nearly 1KHz) to capture activities that are done quickly.
We have observed severely degraded performance of classification methods when sampling rate is around 50 Hz.
Increasing the frame rate increases the number of samples and hence the computational cost increases for de-noising and feature extraction.
Increasing the frame rate may also not help further after some point because human movement speed is limited in indoor areas.
Therefore, by selecting a suitable sampling rate (around 1KHz), a good trade-off between the computational cost and the accuracy can be obtained.

\subsection{Using CSI Phase Information}
Due to the errors such as CFO and SFO, the phase of WiFi CSI has been rarely used for activity recognition in the literature.
However, by subtracting the phase information of neighboring antennas from one another, the CFO and SFO are omitted.
The phase difference is related to the angle of arrival (AOA) although there is integer ambiguity in the number of full cycles of the received signal.
The change in the target location can change the AOA and hence the phase difference.
When the movement is fast and drastic, the signal will be scattered by the human body more randomly, and hence the AOA and phase difference will change faster.
It might be thus helpful to use phase difference together with amplitude for feature extraction and then apply classification algorithms, however, further investigation will be left for future work due to lack of space.

\subsection{Multi-user Activity Recognition}
While the performances of many activity recognition algorithms have been tested for a single user, the more interesting and also challenging problem is the case where multiple people are in the environment.
One solution has been proposed in \cite{Wisee} to use the idea of MIMO receivers to separate the signals due to two distinct mobile objects.
Having multiple receivers might also help in distinguishing the activities of multiple users.
Some techniques for multi-speaker recognition might be applicable to the activity recognition problem, however, this remains an interesting open problem.

\section{Conclusion and Future Work}\label{Sec:Conclusion}
In this work, a survey of recent advancements in human activity recognition systems using WiFi channel has been provided.
The literature in this area shows great promise in achieving good accuracy in indoor environments.
Using numerical test it has been observed that better accuracy can be obtained by employing deep learning techniques such as RNN LSTM rather than methods such as HMM.
There are still several challenges that need to be addressed in future work such as how to use CSI phase information in addition to the amplitude, how to make the system robust in different dynamic environments, and how to identify the behaviours of multiple users.

% or


%\appendices

%\section{}

%\section*{Acknowledgment}
%The authors would like to thank ... of Stanford for the valuable discussion regarding this project. 
%The work is supported in part by the Fund de Researche du Quebec-Nature et technologies (FQRNT).

\ifCLASSOPTIONcaptionsoff
  \newpage
\fi



%\bibliographystyle{IEEEtran}

\bibliographystyle{IEEEtran}
\bibliography{References}


%\vfill


\begin{IEEEbiography}[{\includegraphics[width=1in,height=1.25in,clip,keepaspectratio]{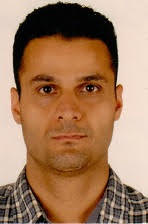}}]{Siamak Yousefi}
received his PhD degree in Electrical and Computer Engineering from McGill University in 2015. Since then he has been a postdoc fellow at the Department of Electrical and Computer Engineering, University of Toronto. He is a recipient of postdoc grant Fonds de recherche du Quebec-nature et technologies (FQRNT). His research interests include applications of statistical signal processing and machine learning techniques for indoor positioning and human activity recognition.
\end{IEEEbiography}

\begin{IEEEbiography}[{\includegraphics[width=1in,height=1.25in,clip,keepaspectratio]{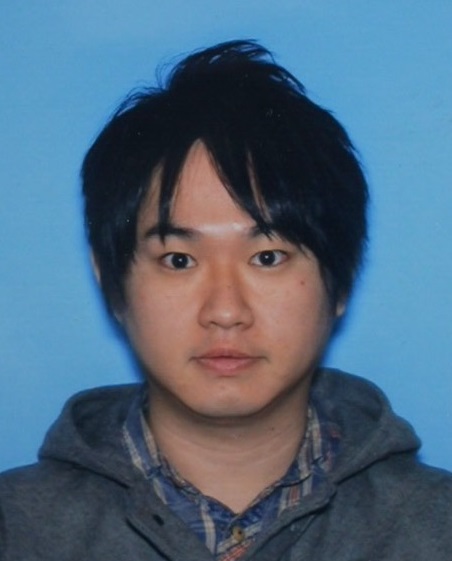}}]{Hirokazu Narui}
received his M.S. degree in Physics and Electronics from Osaka Prefecture University, Japan in 2010. He is currently a visiting researcher in the Department of Computer Science at Stanford University while working at Furukawa Electric Co., Ltd. in Japan. His main research interests are machine learning, neural networks, and indoor positioning.
\end{IEEEbiography}

\begin{IEEEbiography}[{\includegraphics[width=1in,height=1.25in,clip,keepaspectratio]{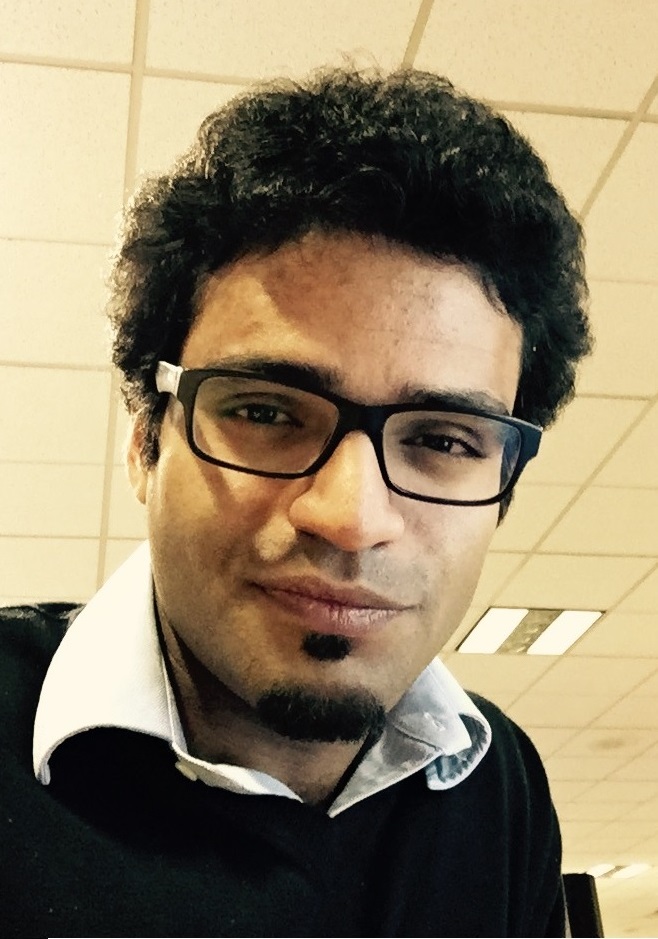}}]{Sankalp Dayal}
received his BS in Electrical Engineering from Indian Institute of Technology, Delhi, and MS in Electrical Engineering from University of California Santa Barbara, and is currently continuing his studies in Artificial Intelligence at Stanford University. His research interests include applied machine learning and signal processing for AI-based consumer applications.
He holds a patent on human motion-based pattern matching. Sankalp is currently working as a Senior Algorithm Engineer in STMicroelectronics.
\end{IEEEbiography}

\begin{IEEEbiography}[{\includegraphics[width=1in,height=1.25in,clip,keepaspectratio]{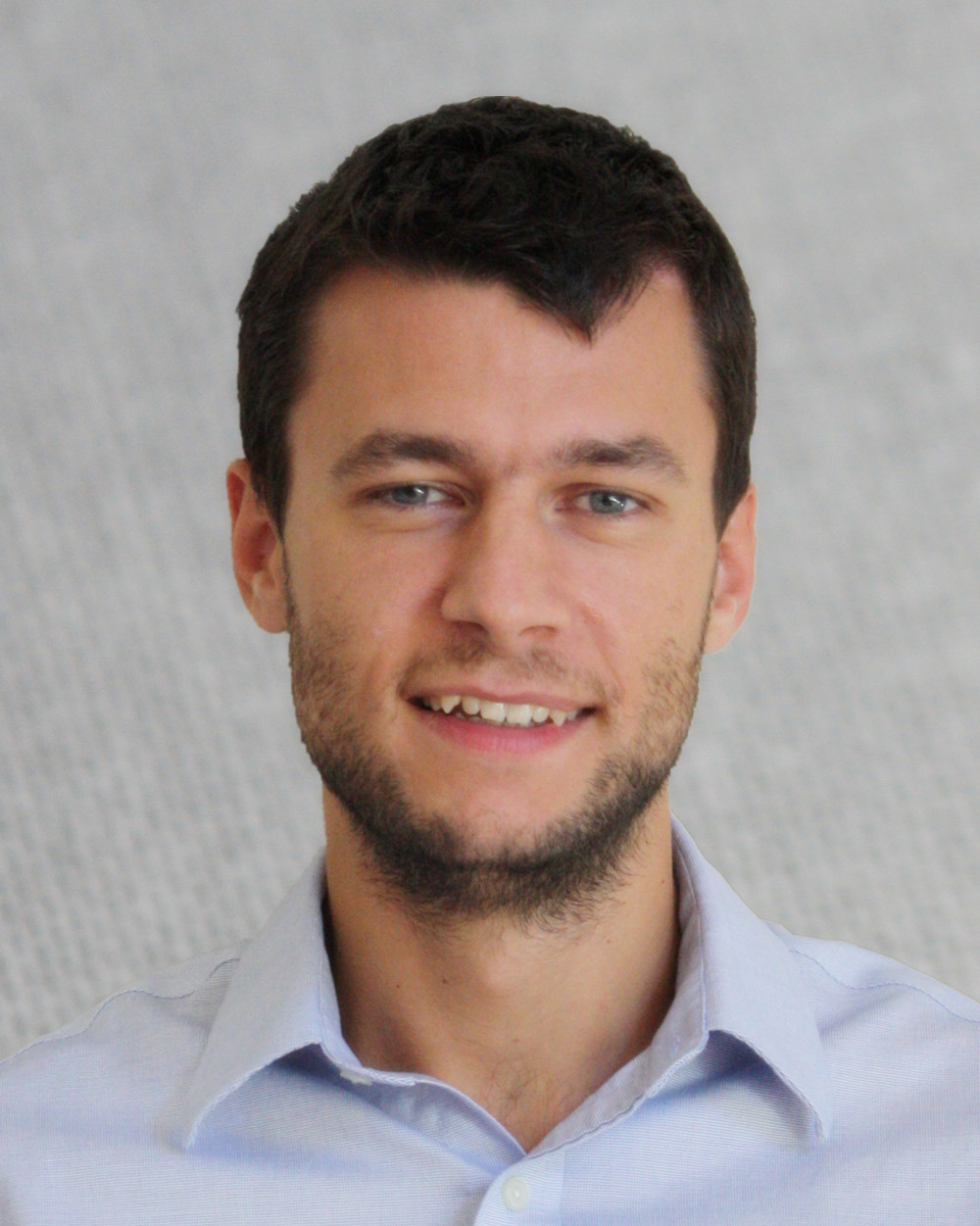}}]{Stefano Ermon}
is an Assistant Professor in the Department of Computer Science at Stanford University, where he is affiliated with the Artificial Intelligence Laboratory and the Woods Institute for the Environment. He received his PhD in Computer Science from Cornell University in 2015. He has co-authored over 40 publications, and has won several best paper awards at AAAI, UAI, and CP. He is a recipient of the NSF Career Award.
\end{IEEEbiography}

\begin{IEEEbiography}[{\includegraphics[width=1in,height=1.25in,clip,keepaspectratio]{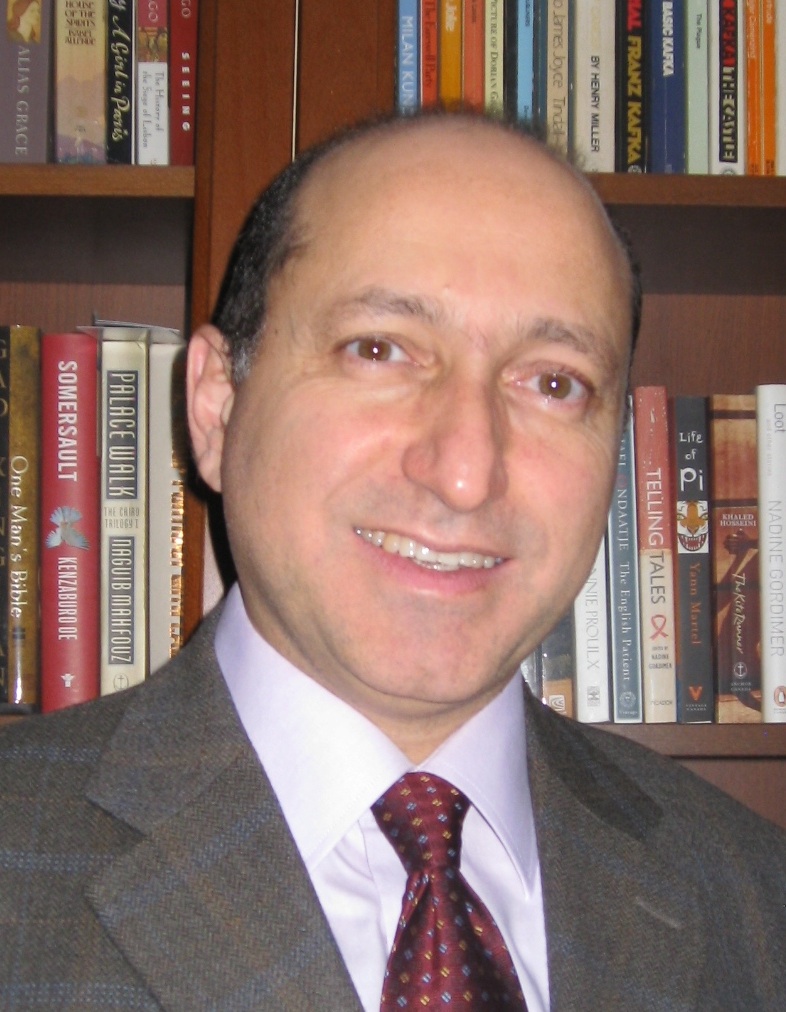}}]{Shahrokh Valaee}
is a Professor with the Department of Electrical and Computer Engineering, University of Toronto, Canada. 
From 2010 to 2012, he was the Associate Editor of IEEE Signal Processing Letters, and from 2010 to 2015 an Editor of IEEE Transactions on Wireless Communications. 
Professor Valaee is a Fellow of the Engineering Institute of Canada.
\end{IEEEbiography}

\end{document}